\documentclass[letterpaper, 10 pt, journal, twoside]{IEEEtran}

\usepackage{graphics} 
\usepackage[export]{adjustbox}  
\usepackage{times} 
\usepackage{cite}
\usepackage{amsmath,amssymb}

\usepackage{multirow}
\usepackage{xspace}

\usepackage{xcolor}
\definecolor{MyLinkColor}{rgb}{0,0.07843,0.45098}
\usepackage[pdftex,colorlinks=true,allcolors=MyLinkColor]{hyperref}

\usepackage{algorithm}
\usepackage{algpseudocode}
\makeatletter
\newcommand{\algmargin}{\the\ALG@thistlm}
\makeatother
\newlength{\whilewidth}
\algdef{SE}[parWHILE]{parWhile}{EndparWhile}[1]
  {\parbox[t]{\dimexpr\linewidth-\algmargin}{%
     \hangindent\whilewidth\strut\algorithmicwhile\ #1\ \algorithmicdo\strut}}{\algorithmicend\ \algorithmicwhile}%
\algnewcommand{\parState}[1]{\State%
  \parbox[t]{\dimexpr\linewidth-\algmargin}{\strut #1\strut}}

\newcommand{\argmin}{\mathop{\mathrm{arg\,min}}}
\newcommand{\paren}[1]{\left(#1\right)}
\newcommand{\tuple}[1]{\left\langle#1\right\rangle}
\newcommand{\R}{\mathbb{R}}
\newcommand{\pardiff}[2]{\frac{\partial #1}{\partial #2}}
\newcommand{\realVectorSpace}[1]{\R^{#1}}
\newcommand{\SE}[1]{\mathrm{SE}(#1)}
\newcommand{\Sphere}[1]{\mathrm{S}(#1)}

\newcommand{\simulator}{f}
\newcommand{\robotState}{\mathbf{x}}
\newcommand{\robotAction}{\mathbf{u}}
\newcommand{\contactWrench}{\mathbf{y}}
\newcommand{\gtContactWrench}{\contactWrench^{\text{gt}}}
\newcommand{\geomParam}{\boldsymbol{\theta}}
\newcommand{\geomParamDim}{P}
\newcommand{\initGeomParam}{\geomParam^{\text{init}}}
\newcommand{\gtGeomParam}{\geomParam^{\text{gt}}}
\newcommand{\estGeomParam}{\hat{\geomParam}}
\newcommand{\goalState}{\robotState^{\text{goal}}}
\newcommand{\goalSet}{\mathcal{G}}
\newcommand{\oracle}{\pi}
\newcommand{\timestep}{t}

\newcommand{\belief}{\mathit{bel}}
\newcommand{\beliefPreObs}{\overline{\mathit{bel}}}
\newcommand{\beliefSize}{N}
\newcommand{\geomParamPar}[2]{\estGeomParam_{#1}^{#2}}
\newcommand{\contactWrenchPar}[2]{\contactWrench_{#1}^{#2}}
\newcommand{\cost}{c}
\newcommand{\costPar}[2]{\cost_{#1}^{#2}}
\newcommand{\historyLen}{H}
\newcommand{\taskDuration}{T}
\newcommand{\rollout}{\mathrm{Rollout}}
\newcommand{\residual}{r}

\newcommand{\gPos}{\mathbf{q}}  
\newcommand{\gVel}{\dot{\mathbf{q}}}  
\newcommand{\gCtrl}{\boldsymbol{\tau}}  
\newcommand{\numContact}{C}  
\newcommand{\collisionFreeFD}{\operatorname{FD}}  
\newcommand{\collisionFD}{\collisionFreeFD_{\text{full}}}  
\newcommand{\toi}{\mathit{toi}}  
\newcommand{\timestepDuration}{\Delta}
\newcommand{\defaultTimestepDuration}{\Delta_{s}}
\newcommand{\contactJacobian}{\mathbf{J}}  
\newcommand{\contactJacobianT}{\contactJacobian^{\top}}  
\newcommand{\contactJacobianTCol}[1]{\contactJacobian^{{#1}^{\top}}}  
\newcommand{\minimumNormalV}{\varepsilon}  
\newcommand{\contactImpulse}{\boldsymbol{\lambda}}  
\newcommand{\lcpMat}{\mathbf{A}}  
\newcommand{\lcpVec}{\mathbf{b}}  
\newcommand{\lcpAcc}{\mathbf{a}}  
\newcommand{\lcp}{\mathrm{LCP}}
\newcommand{\wrenchMap}{\contactJacobianT}
\newcommand{\contactLocation}{\mathbf{p}}
\newcommand{\contactVelocity}{\mathbf{v}}
\newcommand{\contactNormal}{\mathbf{n}}
\newcommand{\firstCollisionObject}{A}
\newcommand{\secondCollisionObject}{B}
\newcommand{\firstContactLocation}{\contactLocation^{\firstCollisionObject}}
\newcommand{\secondContactLocation}{\contactLocation^{\secondCollisionObject}}
\newcommand{\firstContactVelocity}{\contactVelocity^{\firstCollisionObject}}
\newcommand{\secondContactVelocity}{\contactVelocity^{\secondCollisionObject}}
\newcommand{\contactLocationAt}[1]{\contactLocation^{#1}}

\newcommand{\contactNormalAt}[1]{\contactNormal^{#1}}
\newcommand{\contactNormalAtT}[1]{\contactNormal^{{#1}^{\top}}}
\newcommand{\arbitraryVecElement}[1]{z^{#1}}
\newcommand{\arbitraryVec}{\mathbf{z}}

\makeatletter
\DeclareRobustCommand\onedot{\futurelet\@let@token\@onedot}
\def\@onedot{\ifx\@let@token.\else.\null\fi\xspace}
\def\eg{\emph{e.g}\onedot} 
\def\ie{\emph{i.e}\onedot} 
 
\def\etc{\emph{etc}\onedot} \def\vs{\emph{vs}\onedot}
\def\wrt{w.r.t\onedot} 
\def\etal{\emph{et al}\onedot}
\makeatother

\newcommand{\taskGrasp}{\textbf{pose}}

\newcommand{\taskShape}{\textbf{shape}}
\newcommand{\taskPillar}{\textbf{env}}

\newcommand{\secref}[1]{Section~\ref{#1}}
\renewcommand{\eqref}[1]{(\ref{#1})}
\newcommand{\figref}[1]{Fig.~\ref{#1}}

\renewcommand{\algref}[1]{Algorithm~\ref{#1}}

\begin{document}

\title{
Differentiable Contact Dynamics for Stable Object Placement Under Geometric Uncertainties
}

\author{
Linfeng Li$^{1}$,
Gang Yang$^{1}$,
Lin Shao$^{1}$
and David Hsu$^{1,2}$
\thanks{Manuscript received: July 17, 2025; Revised: October 25, 2025; Accepted: November 22, 2025.}
\thanks{This paper was recommended for publication by Editor Clement Gosselin upon evaluation of the Associate Editor and Reviewers' comments.
This research is supported by the National Research Foundation, Singapore under its AI Singapore Programme (AISG Award No: AISG2-PhD-2021-08-014).} 
\thanks{$^{1}$School of Computing, National University of Singapore, Singapore. {\tt\small linfeng@comp.nus.edu.sg}, {\tt\small ygang@u.nus.edu}, {\tt\small linshao@nus.edu.sg}, {\tt\small dyhsu@comp.nus.edu.sg}.
}
\thanks{$^{2}$Smart Systems Institute, National University of Singapore, Singapore.}
\thanks{Digital Object Identifier (DOI): see top of this page.}
}

\markboth{IEEE Robotics and Automation Letters. Preprint Version. Accepted November, 2025}
{Li \MakeLowercase{\textit{et al.}}: Differentiable Contact Dynamics for Object Placement} 

\maketitle

\begin{abstract}
From serving a cup of coffee to positioning mechanical parts during assembly, stable object placement is a crucial skill for future robots. It becomes particularly challenging under geometric uncertainties, \eg, when the object pose or shape is not known accurately. This work leverages a \textit{differentiable simulation model} of contact dynamics to tackle this challenge. We derive a novel gradient that relates force-torque sensor readings to geometric uncertainties, thus enabling uncertainty estimation by minimizing discrepancies between sensor data and model predictions via gradient descent. Gradient-based methods are sensitive to initialization. To mitigate this effect, we maintain a \textit{belief} over multiple estimates and choose the robot action based on the current belief at each timestep. In experiments on a Franka robot arm, our method achieved promising results on multiple objects under various geometric uncertainties, including the in-hand pose uncertainty of a grasped object, the object shape uncertainty, and the environment uncertainty.
\end{abstract}



\section{Introduction}

\IEEEPARstart{F}{uture} robots should play an integral role in aiding humans with everyday tasks. A key capability is to stably place objects in different environments. Such tasks, ranging from serving dishes on a table to carefully rearranging delicate items, might appear simple but are challenging due to the required accuracy. For instance, to avoid spills when serving a full cup of coffee, the robot must place the cup precisely on the saucer without dropping or collision: inaccuracy in a couple of millimeters may cause spilling~(\figref{fig:motivation}). This challenge extends beyond mere motion control, as visual or depth sensors have errors ranging from a few millimeters to centimeters~\cite{realsense_datasheet}, and we rarely know the exact geometries of the object and the environment. Accounting for such geometric uncertainties is critical for robust robotic manipulation.

\begingroup
\setlength{\tabcolsep}{2pt}
\begin{figure}[t]
    \centering
    \includegraphics[scale=0.5]{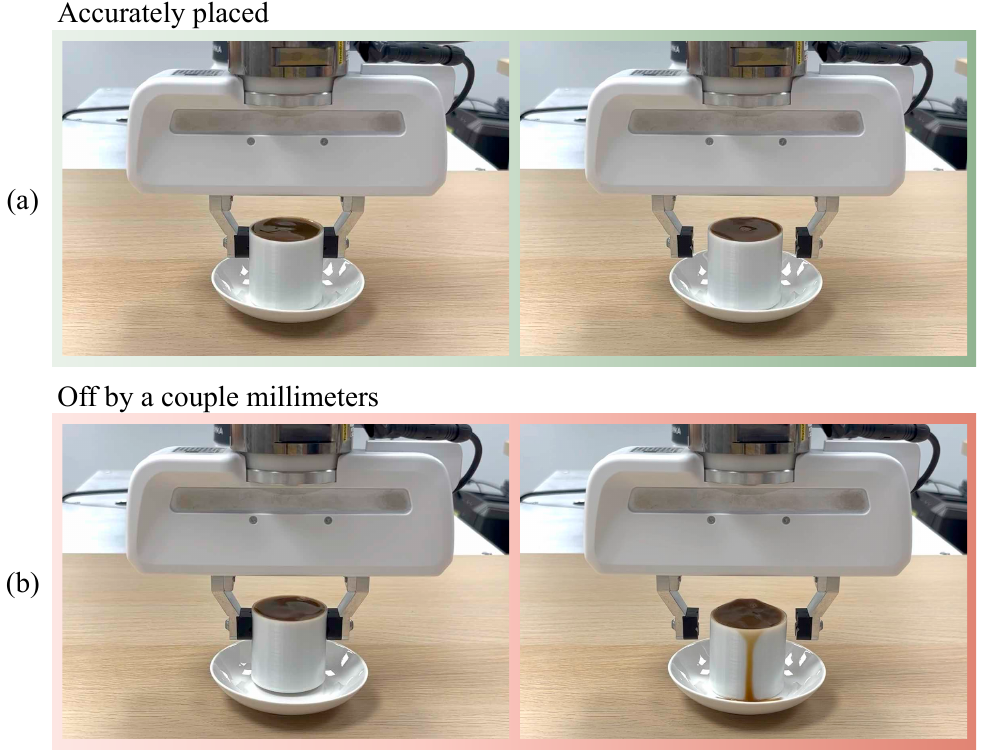}
    \caption{Place a full cup of coffee on a saucer (left) and then release it (right). (a) When the coffee cup is stably placed, \ie, the bottom surface of the cup is well aligned  the saucer, there is no spill after release. (b) If the gripper releases right after a sensed contact without proper estimation of geometric uncertainty, the coffee is spilled.}
    \label{fig:motivation}
\end{figure}
\endgroup

Stable placement is one instance of contact-rich manipulation, where geometric uncertainty has long been a challenge~\cite{rodriguez2021unstable_queen}; exploring stable placement therefore offers insights into this challenge. In contact-rich manipulation, objects interact through the establishment or cessation of contact, leading to hybrid dynamics where each contact mode dictates a unique smooth dynamics. For example, the dynamics of placing a coffee cup changes upon making contact with a saucer, with these dynamics' boundaries shaped by the objects' geometries, such as the saucer's shape. Because of the hybrid dynamics, contact-rich manipulation is sensitive to geometric uncertainty, where “small changes in the geometry of parts can have a significant impact on fine-motion strategies~\cite{lozano-perez1984automatic}.”

Differentiable simulation offers a promising approach to robotic manipulation under uncertainty, as it enables parameter estimation by aligning simulated and measured sensor readings through gradient-based optimization~\cite{li2019learning,mitrano2023focused,le2023differentiable}. However, existing differentiable simulators do not provide gradients of contact wrenches with respect to geometric parameters, and therefore cannot estimate uncertain geometries.

We derive a novel gradient that relates the predicted force-torque readings to the geometric parameters, a feature underexplored in existing differentiable rigid-body simulators. This gradient enables the estimation of uncertain geometry by aligning simulated and measured force–torque data through gradient descent. We implemented this feature on top of the Jade simulator~\cite{yang2024jade}, whose generality allows us to address a broad range of geometric uncertainties. We verify the efficacy of the derived gradient using the stable object placement task. To mitigate the sensitivity of gradient-based methods to initialization, we maintain a belief over multiple geometric parameters. At each timestep, we update the belief using the recent history, and choose the robot action based on the current belief. We evaluated our method with a Franka Research~3 robot arm under various geometric uncertainties, including in-hand pose uncertainty of the grasped object, the object’s shape uncertainty, and the environment’s shape uncertainty. Our method achieved promising results, compared with a trigger-based heuristic policy and a particle filter--based policy. Finally, we demonstrate our method’s practicality by using it to place a full cup of coffee.

This paper primarily focuses on the geometric uncertainty. Other forms of uncertainty, such as sensor inaccuracies or control imprecision, are addressed through calibration.

\section{Related Work}
\label{sec:related-work}

\subsection{Robotic Stable Placement}

There are various studies on robotic stable placement with haptic feedback~\cite{jain2010el_e,romano2011human,kim2023simultaneous,ota2024tactile}. However, existing work often assumes a specific type of geometric uncertainty: the object will be placed on a large flat surface~\cite{jain2010el_e,romano2011human}, the initial contact between the object and the surface has to be a point contact with upward contact normal~\cite{kim2023simultaneous}, the test cases are similar to the training data~\cite{ota2024tactile}, \etc. Our method applies to a broader range of geometric uncertainties, including the uncertainty in the grasped object's in-hand pose, the object's shape, and the environment's shape. These three types cover the geometric uncertainties considered in prior works, making our approach more general and adaptable.

\subsection{Manipulation Under Geometric Uncertainty}

Geometric uncertainty has long been a challenge in manipulation \cite{rodriguez2021unstable_queen}. For a particular setup, a manipulation policy can be manually crafted~\cite{romano2011human,stueckler2014adaptive,suarez-ruiz2018can} or learned~\cite{inoue2017deep,johannink2019residual,jin2021contact,ota2024tactile,zhang2024bridging}; nevertheless, their generalizations are limited by the data diversity or the assumptions that guide the policy-crafting. For example, in a learning-based approach for stable placement~\cite{ota2024tactile}, the object's bottom surface and the environmental support surface are assumed to be flat and parallel. Our approach alleviates these restrictions by following the model-based approach.

Model-based methods hold the promise of generalizing to different setups. For simpler tasks, a robust sequence of actions can be planned to achieve the goal under all possible variations of the uncertain geometry~\cite{erdmann1988exploration,goldberg1993orienting,lozano-perez1984automatic,wirnshofer2018robust,chintalapudi2024bi}. Alternatively, the problem can be modeled as a partially observable Markov decision process~\cite{hsiao2007grasping_pomdps,koval2016pre,saund2023blindfolded}. However, these methods only utilize binary haptic sensor feedback (touch \vs no-touch) or simple termination conditions (\eg, y-force larger than a threshold); this limits their applicability to more complex tasks.

Many prior works estimate uncertain geometry using predefined probing procedures, where robot actions follow constrained motion patterns, such as following a fixed trajectory or choosing a location on a plane to move downwards until a force threshold is met~\cite{chhatpar2005localization,taguchi2010rao,drigalski2021precise,saund2022clasp,ma2021extrinsic,gadeyne2005bayesian,hertkorn2012identification,sipos2022simultaneous,sipos2023multiscope,lee2023uncertain,thomas2007multi}. These probing procedures are tailored to specific types of uncertainties, limiting their generalization, and the quality of estimation heavily depends on their design.

By using a general differentiable contact dynamics model, we estimate various geometric uncertainties in a unified way: minimizing the discrepancy between the model prediction and the multi-dimensional continuous-value haptic sensor readings.

\subsection{Differentiable Contact Dynamics}

Differentiable physical models have been shown to be useful in estimating the uncertain parameters~\cite{li2019learning,mitrano2023focused,le2023differentiable}; the discrepancy between the model prediction and the observation is minimized using the gradients. We take a similar approach in the context of rigid-body contact dynamics.

To estimate the uncertain geometry, we need the gradients of contact forces with respect to the uncertain geometric parameters. Prior differentiable rigid-body simulators do not provide this capability~\cite{avila2018end,heiden2021neuralsim,degrave2019differentiable,qiao2020scalable,qiao2021efficient,werling2021fast,geilinger2020add,howell2022dojo}. Lee~\etal propose a contact feature that is differentiable \wrt the object pose~\cite{lee2023uncertain}; a differentiable contact feature is an important intermediary step, but the feature itself does not suffice to predict the contact force. We derived and implemented the gradients of contact forces with respect to the uncertain geometric parameters on top of the Jade simulator~\cite{yang2024jade}.

\section{Problem Formulation}
\label{sec:formulation}

Given an initial guess of the uncertain geometry, our method takes the robot’s proprioception and the 6-axis force-torque (FT) sensor feedback as input; it outputs the robot action to approach the goal configuration to stably release the target object.

We consider three entities in our model: the end-effector frame, the object to be placed, and the environment. The object and the environment are represented as triangle meshes. We model them using a differentiable simulator
\begin{equation}
  \robotState_{t+1},~ \contactWrench_{{t+1}} = \simulator(\robotState_{t}, \robotAction_{t}; \geomParam),
  \label{eqt:simulator}
\end{equation}
where $\robotState_{t}\in \SE{3} \times \realVectorSpace{6}$ is the robot state consisting of the end-effector's pose and twist, $\robotAction_{t}\in \SE{3}$ is the robot action of reference pose, which is converted into a control wrench by a lower-level impedance controller, and $\contactWrench_{t}\in \realVectorSpace{6}$ is the predicted contact wrench between the object and the environment expressed in the end-effector frame. We always use the subscript~$t$ in~$(\cdot)_{t}$ to denote the timestep. The uncertain geometry is represented by the parameter $\geomParam \in \realVectorSpace{\geomParamDim}$, which includes parameters of i) the relative pose between the object and the end-effector, ii) the mesh coordinates of the object, and iii) the mesh coordinates of the environment. The users have the flexibility to define the elements of~$\geomParam$ based on their specific placement task requirements. The dynamics function $\simulator$ is differentiable with respect to $\robotState_{t}$, $\robotAction_{t}$, and $\geomParam$.

After the robot grasps the target object, we start with an initial guess $\initGeomParam \in \realVectorSpace{\geomParamDim}$ of the unknown groundtruth geometry $\gtGeomParam \in \realVectorSpace{\geomParamDim}$. We assume $\gtGeomParam$ to be a constant. For example, this assumption implies a non-slip grasp, ensuring that the uncertain in-hand pose of the object does not change during execution.

To stably place the grasped object, the robot needs to release the object at a goal
\begin{equation}
\goalState \in \goalSet(\gtGeomParam)
\label{eqt:goal}
\end{equation}
determined by the unknown groundtruth $\gtGeomParam$, where $\goalSet$ maps a geometric parameter to a set of goal states. For example, to place a cup on a table,~$\goalSet(\gtGeomParam)$ includes all configurations where bottoms of the cup are aligned with the table surface.

We assume a deterministic policy
\begin{equation}
 \robotAction_{t} = \oracle(\robotState_{t}, \geomParam)
\label{eqt:oracle-policy}
\end{equation}
that takes the robot state and geometric parameter as inputs and outputs the robot action $\robotAction_{t}$. When the groundtruth geometry is used, this policy drives the robot to the stable release configuration, \ie, $\lim_{t\to \infty}\robotState_{t} \in \goalSet(\gtGeomParam)$ for the closed-loop dynamics $ \robotState_{t+1},~ \contactWrench_{t+1} = \simulator\left(\robotState_{t}, \oracle(\robotState_{t}, \gtGeomParam); \gtGeomParam\right)$. This assumption reflects that when the geometry is known, we can leverage existing methods for stable placement. In general, this policy can be implemented using motion planning methods~\cite{toussaint2018differentiable,pang2023global,le2024fast}, or learned conditioned on the groundtruth geometry~\cite{kumar2021rma}. In this work, we use heuristic policies that interpolate from the current state to the goal.

Our method relies on the FT sensor to measure the contact wrench between the object and the environment $\gtContactWrench_{t} \in \realVectorSpace{6}$ at timestep $t$. By minimizing the error between the prediction $\contactWrench_{t}$ and the measurement $\gtContactWrench_{t}$ through gradient descent, we expect the estimated geometric parameter $\estGeomParam$ to approach $\gtGeomParam$, and $\oracle(\robotState_{t}, \estGeomParam)$ drives the robot to the stable release configuration.

We assume uncertainties other than the geometric ones are compensated for by calibration.

\section{Gradients of Geometric Parameters}

To differentiate the error between the model prediction~$\contactWrench_{t}$ and the measurement~$\gtContactWrench_{t}$ with respect to the geometric parameter~$\geomParam$, we require the gradient $\partial \contactWrench_{(\cdot)} / \partial \geomParam$ from the simulator~\eqref{eqt:simulator}. However, this feature is absent in most existing differentiable rigid-body simulators. To address this limitation, we extend the Jade simulator to support differentiation with respect to geometric parameters~\cite{yang2024jade}. Our key design choice is to model contact features not only as functions of configuration but also as functions of the geometric parameter. In this section, we detail the derivation and implementation of this gradient.

\subsection{Robot States and Actions}
\label{sec:robot-states-and-actions}

The robot state $\robotState_{t} = \langle \gPos_{t}, \gVel_{t}\rangle$ consists of the pose $\gPos_{t}\in\SE{3}$ and twist $\gVel_{t}\in \realVectorSpace{6}$ of the end-effector. We model the end-effector and grasped object as a floating rigid body controlled by a wrench~$\gCtrl_t\in \realVectorSpace{6}$, which is computed from the robot action~$\robotAction_{t}$. For example, if $\robotAction_{t}$ is the reference pose of a proportional controller~\cite{astroem2010feedback_systems}, the control wrench is given as~$\gCtrl_{t} = \textit{ControllerGain} \cdot \mathrm{PoseError} (\robotAction_{t}, \gPos_{t})$, where $\mathrm{PoseError}$ computes the 6-dimensional pose difference vector between two poses~\cite{lynch2017modern, murray1995proportional}.

\subsection{Collision-Free Forward Dynamics}
\label{sec:collision-free-forward-dynamics}

We start with the simple case where no contact change occurs during each forward simulation step. We represent contacts as points between object pairs. Given $\gPos$ and $\geomParam$, the collision engine detects all contact features, each as a tuple with respect to objects $\firstCollisionObject$ and $\secondCollisionObject$
\begin{equation}
  \label{eqt:contact-feature}
  \tuple{
    \firstContactLocation(\gPos; \geomParam),~
    \secondContactLocation(\gPos; \geomParam),~
    \firstContactVelocity,~
    \secondContactVelocity,~
    \contactNormal(\gPos; \geomParam)
  }_i,
\end{equation}
where the subscript $i$ denotes the $i$-th contact feature,~$\firstContactLocation, \secondContactLocation \in \realVectorSpace{3}$ are the coordinates of witness points on objects $\firstCollisionObject$ and $\secondCollisionObject$,~$\firstContactVelocity, \secondContactVelocity \in \realVectorSpace{3}$ are the velocities of witness points, and~$\contactNormal \in \Sphere{3}$ is the unit normal vector of contact surface pointing from object $\firstCollisionObject$ to object $\secondCollisionObject$. When the collision happens,  we have $\firstContactLocation = \secondContactLocation$, meaning the witness points coincide and become contact points.

Given the detected $\numContact$ contact features \eqref{eqt:contact-feature}, the simulator predicts the position~$\gPos'$, velocity~$\gVel'$ and contact impulse~$\contactImpulse\in\realVectorSpace{3\numContact}$ after applying the control force $\gCtrl$ for the time duration~$\timestepDuration$ using the forward dynamics
\begin{equation}
  \label{eqt:collision-free-forward}
  \gPos',~\gVel',~ \contactImpulse = \collisionFreeFD\paren{\gPos, \gVel, \gCtrl, \timestepDuration;\geomParam}
\end{equation}
under the assumption that the contact features do not change during $\timestepDuration$, \ie, there is no collision during $\timestepDuration$. Let $\defaultTimestepDuration$ be the duration between two timesteps $\timestep$ and $\timestep + 1$, we have $\gPos_{\timestep+1},\gVel_{t+1},\contactImpulse_{\timestep}=\collisionFreeFD(\gPos_{\timestep},\gVel_{\timestep},\gCtrl_{\timestep},\defaultTimestepDuration; \geomParam)$, and the predicted contact wrench is $\contactWrench_{\timestep+1} = \contactJacobianT(\gPos_{\timestep}; \geomParam) \contactImpulse_{\timestep}$, where $\contactJacobian\in\realVectorSpace{6\times 3\numContact}$ is the contact Jacobian that depends on \eqref{eqt:contact-feature}.

We solve for~$\contactImpulse$ in~\eqref{eqt:collision-free-forward} via a linear complementarity problem $\contactImpulse = \lcp\paren{\lcpMat, \lcpVec}$: finding $ \langle \contactImpulse, ~\lcpAcc \rangle $ such that $\contactImpulse \geq 0$, $\lcpAcc\geq 0$, $\contactImpulse^{\top} \lcpAcc = 0$, and $\lcpAcc = \lcpMat \contactImpulse + \lcpVec$~\cite{werling2021fast, yang2024jade, cottle2009linear}. Here we highlight that the $\lcp$ parameters $\lcpMat=\lcpMat(\contactJacobian)$ and $\lcpVec=\lcpVec(\contactJacobian, \timestepDuration)$ depend on $\contactJacobian$ and~$\timestepDuration$.

However, the assumption of no contact change during the forward simulation step~\eqref{eqt:collision-free-forward} rarely holds. We alleviate this assumption in the next subsection.

\subsection{Collision Between Two Timesteps}
\label{sec:continuous-collision}

If there is a collision between timesteps~$\timestep$ and ~$\timestep+1$ that changes the contact features~\eqref{eqt:contact-feature}, we can detect the exact time-of-impact $\toi\in[0, 1)$ as a ratio of $\defaultTimestepDuration$ when two rigid bodies collide. The general forward dynamics
\begin{equation}
  \label{eqt:collision-forward}
  \gPos_{t+1},~\gVel_{t+1} = \collisionFD\paren{
    \gPos_{t}, \gVel_{t}, \gCtrl_{t}, \defaultTimestepDuration;\geomParam
  }
\end{equation}
might involve multiple calls of collision-free forward dynamics~\cite{yang2024jade}. In the case of 2 calls, we have
\begin{equation*}
  \begin{aligned}
    \gPos_{t+\toi},~\gVel_{t+\toi},~\contactImpulse_{t}
      &= \collisionFreeFD\paren{
        \gPos_{t}, \gVel_{t}, \gCtrl_{t},
        \toi\defaultTimestepDuration;\geomParam
      }, \\
    \gPos_{t+1},~\gVel_{t+1},~\contactImpulse_{t+\toi}
      &= \\[-2pt]
      \collisionFreeFD(
        \gPos_{t+\toi}, &\gVel_{t+\toi},~\gCtrl_{t},
        (1-\toi)\defaultTimestepDuration;\geomParam
      ).
  \end{aligned}
\end{equation*}
We have so far defined the forward dynamics of the simulator~\eqref{eqt:simulator}. Prior works~\cite{werling2021fast, yang2024jade} have implemented the gradients~$\partial\collisionFreeFD / \partial\gPos$,~$\partial\collisionFreeFD / \partial\gVel$,~$\partial\collisionFreeFD / \partial\gCtrl$, and~$\partial\collisionFreeFD / \partial\timestepDuration$. However, our method needs~$\partial\contactWrench_{(\cdot)} / \partial \geomParam$ from the simulator~\eqref{eqt:simulator}. We elaborate on its derivation in the next subsection.

\subsection{Differentiating Wrench wrt Geometric Parameters}
\label{sec:geometric-grad-derivation}

Since in our problem setup, $\gPos_{t}$ is the pose of the robot's end-effector, the contact wrench impulse between the grasped object and the environment is
given as
\begin{equation}
  \label{eqt:predicted-wrench}
  \contactWrench_{t+1} = \left\{
    \begin{array}{ll}
      \wrenchMap_{t+\toi} \contactImpulse_{t+\toi} +
      \wrenchMap_{t} \contactImpulse_{t} & \text{ collision,}\\
      \wrenchMap_{t} \contactImpulse_{t} & \text{ no collision.}
    \end{array}
  \right.
\end{equation}
By applying the chain rule, the gradient of the geometric parameter $\partial \contactWrench_{t+1} / \partial \geomParam$ involves the differentiation of contact impulses $\partial \contactImpulse_{(\cdot)} / \partial \geomParam$ and the Jacobian $\partial \wrenchMap_{(\cdot)} / \partial \geomParam$, where $\partial \contactImpulse_{(\cdot)} / \partial \geomParam$ involves $\partial \contactJacobianT_{(\cdot)} / \partial \geomParam$, $\partial \contactJacobian_{(\cdot)} / \partial \geomParam$ and $\partial \toi / \partial \geomParam$.

To derive $\partial \toi / \partial \geomParam$, we consider the dynamics at timestep~$t+\toi$ as discussed in~\cite{yang2024jade}. Suppose a tiny geometry variant~$\delta\geomParam$ causes coordinate variants~$\delta\contactLocation^A, \delta\contactLocation^B$ and makes the contact points separate along the normal direction, it'll take
\begin{equation*}
    \delta\toi = \frac{(\delta\secondContactLocation - \delta\firstContactLocation) \cdot \contactNormal}{(\secondContactVelocity - \firstContactVelocity) \cdot \contactNormal}
\end{equation*}
longer for them to collide and vice versa. Denote~$v_n = (\secondContactVelocity - \firstContactVelocity) \cdot \contactNormal$ as the relative normal velocity. We have
\begin{equation}
  \pardiff{\toi}{\geomParam} =
  \left.\frac{\contactNormal^{\top}}{\mathrm{clamp}(v_{n},~\minimumNormalV)} \left(\pardiff{\secondContactLocation}{\geomParam} - \pardiff{\firstContactLocation}{\geomParam}\right)\right|_{\gPos=\gPos_{t+\toi}},
\end{equation}
where $\minimumNormalV > 0$ is a small positive constant for clamping to avoid division by zero. When there is no collision, we treat $\timestepDuration$ as a constant in~\eqref{eqt:collision-forward}.

The gradients of Jacobian $\partial \contactJacobianT / \partial \geomParam$ and $\partial \contactJacobian / \partial \geomParam$ are always calculated in the context of $\partial \left( \contactJacobianT \arbitraryVec \right) / \partial \geomParam$ and $\partial \left( \contactJacobian \arbitraryVec' \right) / \partial \geomParam$, with $\arbitraryVec$ and $\arbitraryVec'$ being arbitrary constant vectors. In our setting, the $i$-th column $\contactJacobianTCol{i} = \left[ (\contactNormalAt{i}\times \contactLocationAt{i})^{\top}~ \contactNormalAtT{i} \right]^{\top}$ of $\contactJacobianT$ is defined by the location $\contactLocationAt{i}$ and normal $\contactNormalAt{i}$ of the corresponding contact feature~\eqref{eqt:contact-feature}. Then, $\contactJacobianT \arbitraryVec = \sum_i \contactJacobianTCol{i} \arbitraryVecElement{i}$  and
\begin{equation*}
\pardiff{\left( \contactJacobianT \arbitraryVec \right)}{\geomParam} =
  \sum_i 
  \left[
    \begin{array}{c}
      \left(\partial \contactNormalAt{i} / \partial \geomParam\right) \times\contactLocationAt{i} +
      \contactNormalAt{i} \times \left(\partial \contactLocationAt{i} / \partial \geomParam\right) \\
      \left(\partial \contactNormalAt{i} / \partial \geomParam\right)
    \end{array}
  \right] \arbitraryVecElement{i},
\end{equation*}
where $\arbitraryVecElement{i} \in \realVectorSpace{}$ is the $i$-th element of $\arbitraryVec$. We derive $\partial \left( \contactJacobian \arbitraryVec' \right) / \partial \geomParam$ in a similar manner.

For our current implementation, the user needs to specify the callback functions to calculate the derivatives of the contact features $\contactLocation_{(\cdot)}$ and $\contactNormal_{(\cdot)}$ with respect to the geometric parameter $\geomParam$ for each URDF file. We expect future simulators to provide automatic routines in the collision engine.

\begin{figure}[t]
    \centering
    \includegraphics[scale=1]{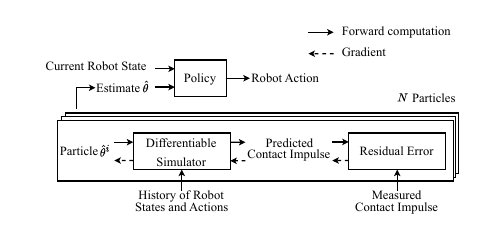}
    \caption{Our overall pipeline. At every timestep, we update the estimated geometry and plan a robot action from the estimation.}
    \label{fig:overall-pipeline}
\end{figure}

\section{Stable Object Placement}
\label{sec:pipeline}

In this section, we demonstrate how our newly derived gradients can address geometric uncertainties in stable object placement. At each timestep, we use gradient descent to refine the estimate of the uncertain geometry and update the robot’s action accordingly. This process repeats for $\taskDuration$ steps before the robot releases the object. Throughout execution, the estimate approaches the groundtruth geometry, guiding the robot toward a stable placement.

\subsection{Maintaining A Single Estimate}
\label{sec:estimating-uncertain-geometry}

We begin with the case of maintaining a single estimate~$\estGeomParam_{t}$, initialized as $\estGeomParam_{0} = \initGeomParam$. At each timestep $t$, we update the estimate $\estGeomParam_{t-1}$ to $\estGeomParam_{t}$ using data from the past $\historyLen + 1$ timesteps, including the robot states $\robotState_{t-\historyLen:t}$, robot actions $\robotAction_{t-\historyLen:t}$, and measured contact wrench $\gtContactWrench_{t-\historyLen+1:t}$. Specifically, we refine the estimate by minimizing the discrepancy between the predicted contact wrench $\contactWrench_{t-H+1:t}$ and the force-torque (FT) sensor measurements $\gtContactWrench_{t-H+1:t}$.

Given a geometric parameter $\geomParam$, we use the history data to get the predicted contact wrench
\begin{equation}
  \label{eqt:rollout}
  \contactWrenchPar{t - \historyLen + 1:t}{} = \rollout\left(
    \robotState_{t - \historyLen : t - 1}, \robotAction_{t - \historyLen : t - 1},\geomParam
  \right),
\end{equation}
where $\rollout$ is defined by applying the simulator~\eqref{eqt:simulator} iteratively: $\left(\cdot\right),~ \contactWrenchPar{t'+1}{} = f(x_{t'}, u_{t'}; \geomParam)$ for $t' = t-H,~ t-H+1,~ \dots,~ t-1$.

We define the discrepancy as the residual error between the measurement and the prediction
\begin{equation}
  \label{eqt:residual-error}
  \residual_{t}(\geomParam) :=
  \residual\left(\gtContactWrench_{t-H+1:t},~ \contactWrench_{t-H+1:t} \right)
  \in \realVectorSpace{},
\end{equation}
where $\residual\geq 0$ measures the distance between two sequences of contact wrenches; when the two sequences are identical, the residual is zero.

We find the updated geometric parameter that minimizes the residual error $\estGeomParam_{t} = \argmin_{\geomParam}\; \residual_{t}(\geomParam)$ by gradient descents from $\estGeomParam_{t-1}$. We summarize the estimation procedure in \algref{alg:geometry-update}.

Then, we compute the robot action using the policy $\robotAction_{t}=\oracle\left(\robotState_{t}, \estGeomParam_{t}\right)$. Finally, we send $\robotAction_{t}$ to the robot for execution.

\begin{algorithm}[t]
\small
  \caption{\small
    Procedure to update a single geometric parameter and its cost using the history.
  }
  \label{alg:geometry-update}
  \begin{algorithmic}[1]
    \Procedure{Geometry-Update}
    {
      $\estGeomParam_{t-1}$,
      $\robotState_{t - \historyLen : t}$,
      $\robotAction_{t - \historyLen : t}$,
      $\gtContactWrench_{t - \historyLen + 1 : t}$
    }
    \State
    Initialize:
    $\geomParam \leftarrow \estGeomParam_{t-1}$.
    \For{number of iterations}
    \State Compute the residual $\residual_{t}(\geomParam)$ defined in (\ref{eqt:residual-error}).
    \State $\textit{gradient} \leftarrow \partial\residual_{t} / \partial\geomParam$.
    \If{$\textit{gradient} = 0$}
    \parState
    {%
      \Return $ \estGeomParam_{t} = \geomParam $, $\cost = \residual_{t}(\theta)$.
    }
    \EndIf
    \parState
    {%
      $\geomParam \leftarrow \geomParam - \textit{learningRate} \cdot \mathrm{clipGrad}(\textit{gradient})$. \label{line:geometry-update}
    }
    \EndFor
    \State \Return $ \estGeomParam_{t} = \geomParam $, $\cost = \residual_{t}(\theta)$.
    \EndProcedure
  \end{algorithmic}
\end{algorithm}

{
\setlength{\textfloatsep}{2pt}
\setlength{\intextsep}{0pt}  
\begin{algorithm}[t]
\small
  \caption{\small
    Procedure to update the belief using the history and our gradient-based estimator.
  }
  \label{alg:belief-update}
  \begin{algorithmic}[1]
    \Procedure{Belief-Update}{
      $\belief_{t-1}$,
      $\robotState_{t - \historyLen : t}$,
      $\robotAction_{t - \historyLen : t}$,
      $\gtContactWrench_{t - \historyLen + 1 : t}$
    }
    \State
    Initialize:
    $\belief_{t} \leftarrow \varnothing$.
    \For{parameter $\geomParam$ in $\belief_{t-1}$}
    \parState
    {%
      $\textit{updatedParticle} \leftarrow$ \Call{Geometry-Update}{
        $\geomParam$,
        $\robotState_{t - \historyLen : t}$,
        $\robotAction_{t - \historyLen : t}$,
        $\gtContactWrench_{t - \historyLen + 1 : t}$}
    }
    \parState
    {%
      $\belief_{t} \leftarrow \belief_{t} \cup \left\{\textit{updatedParticle}\right\} $
    }
    \EndFor
    \State \Return $\belief_{t}$.
    \EndProcedure
  \end{algorithmic}
\end{algorithm}
}

\begingroup
\setlength{\tabcolsep}{2pt}
\begin{figure*}[t]
    \centering
    \includegraphics[scale=0.5]{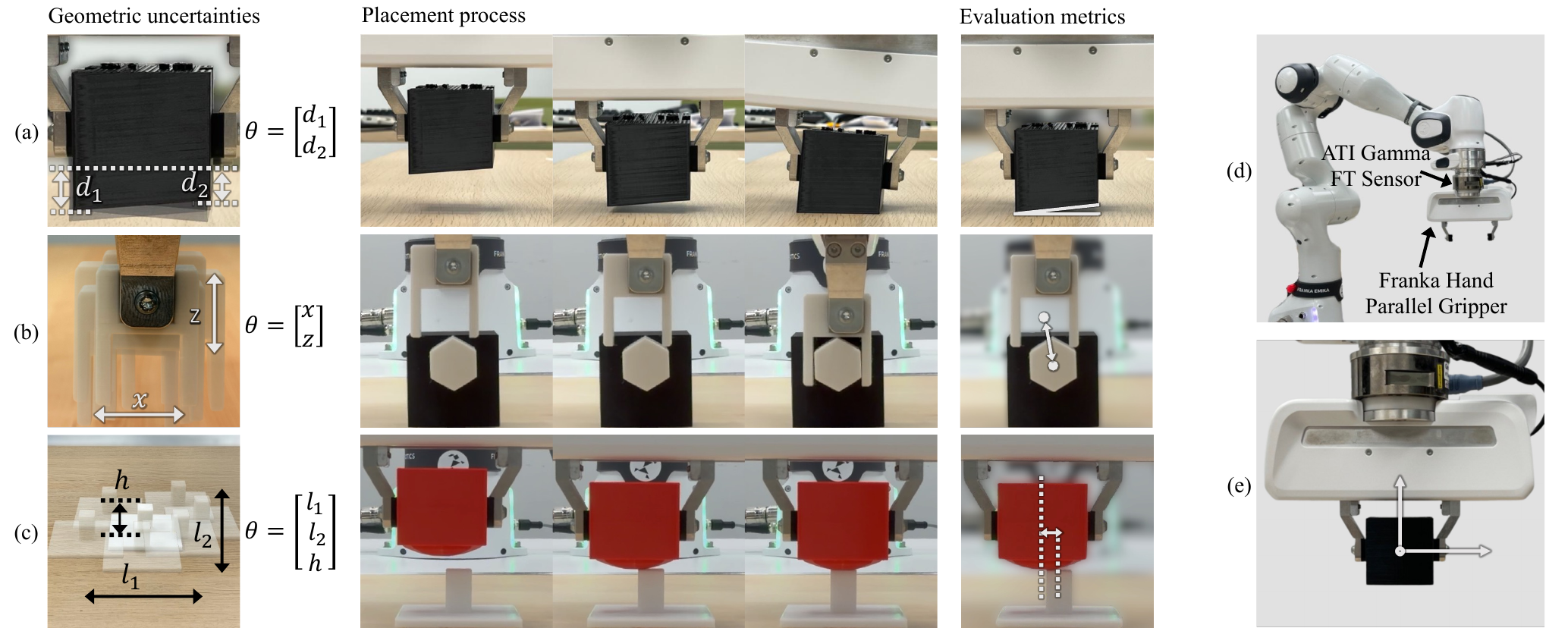}
    \caption{Experimental setups. (a)-(c) Visualization of the task setups for (a) \taskShape, (b) \taskGrasp\ and (c) \taskPillar: the left column shows the unknown geometric parameters, the middle columns show an ideal placement process, and the right column visualizes the distance to goal when releasing. (d) The ATI Gamma sensor and Franka Hand mounted on a Franka Research 3 arm. (e) The end-effector frame in which the motion and forces are represented.}
    \label{fig:experimental-setup}
\end{figure*}
\endgroup

\subsection{Multiple Initializations}
\label{sec:multiple-init}

Gradient descents only find the local minimum and are sensitive to the initialization. To mitigate this, we keep track of a belief over multiple geometric parameters over time.

We represent the belief $\belief_{t}$ as a set of $\beliefSize$ particles, with each particle being a tuple of the value and the cost of the parameter
\begin{align}
  \belief_{t} = \left\{
    \tuple{\geomParamPar{t}{1},~\costPar{t}{1}},~
    \tuple{\geomParamPar{t}{2},~\costPar{t}{2}},~
    \dots,~
    \tuple{\geomParamPar{t}{N},~\costPar{t}{\beliefSize}}
  \right\},
  \label{eqt:belief}
\end{align}
where $\geomParamPar{\timestep}{i}$ is the $i$-th parameter at timestep $\timestep$, and
$ \costPar{\timestep}{i} = \residual_{\timestep}\paren{\geomParamPar{\timestep}{i}} \geq 0 $
is the cost of that parameter.
To initialize the belief $\belief_{{0}}$, we set the value $\geomParamPar{0}{i} = \initGeomParam + \epsilon$ with $\epsilon$ sampled from a distribution of the perception noise, and the cost $\costPar{0}{i}$ to be an arbitrary positive constant, for $i=1,~2,~\dots,~\beliefSize$.

At every timestep $t$, we first update the belief from $\belief_{t-1}$ to $\belief_{t}$ using the history of the past $\historyLen + 1$ timesteps, including the states $\robotState_{t-\historyLen:t}$, actions $\robotAction_{t-\historyLen : t}$, and measurements $\gtContactWrench_{t-\historyLen+1:t}$ (see \algref{alg:belief-update}). Then, we select the value of the particle with the minimum cost as the estimate $\estGeomParam_{t}$, and compute the robot action using the policy $\robotAction_{t}=\oracle(\robotState_{t}, \estGeomParam_{t})$. Finally, we send $\robotAction_{t}$ to the robot for execution. We repeat this process for $\taskDuration$ steps before the robot releases the object. The overall pipeline is illustrated in~\figref{fig:overall-pipeline}.

\section{Experiment}

\subsection{Tasks}

We evaluate our proposed method on 3 tasks with different kinds of geometric uncertainty: \taskShape, \taskGrasp, and \taskPillar. The geometric uncertainty and placement processes of the 3 tasks are illustrated in~\figref{fig:experimental-setup}.

In \taskShape, the robot needs to stably place a cube on the table surface. The goal set $\goalSet$ in~\eqref{eqt:goal} is defined by all configurations in which the bottom surface of the cube is aligned with the table surface. The cube's left and right walls have uncertain heights, defined by the deviations from the nominal values~$d_{1}$ and~$d_{2}$.

In \taskGrasp, the robot needs to stably place a wrench on a hex-shaped part. The goal set $\goalSet$ in~\eqref{eqt:goal} contains a single configuration in which the wrench is aligned with the part. The wrench's in-hand pose has uncertainty, represented as the deviations $x$ and $z$.

In \taskPillar, the robot needs to place a block with a curved bottom surface on a pillar, whose top square surface is parallel to the table surface; the square's side length is 15 mm. The goal set $\goalSet$ in~\eqref{eqt:goal} contains a single configuration in which the center of the object's bottom surface coincides with that of the pillar's top surface. The horizontal locations and height of the pillar have uncertainty, represented as the deviations from the nominal values~$l_1$,~$l_2$ and~$h$.

In all tasks, the nominal values of the uncertain geometry are $0$, and are used as the initial guess $\initGeomParam = 0$. Each task has 10 test cases of different groundtruth geometric parameters $\gtGeomParam \neq \initGeomParam$. The groundtruth geometric parameters $\gtGeomParam$ are obtained from manual calibration and CAD models of the 3D-printed parts. We set the duration $\taskDuration = 50$ and the history $\historyLen = 5$ timesteps for all tasks. At the initial robot configuration, the gripper is above a \emph{nominal} goal defined by the \emph{nominal} geometric parameter $\goalSet(\initGeomParam)$.

\subsection{Setup}

We carry out the experiments on a Franka Research 3 (FR3) arm. We mount an ATI Gamma FT sensor on the flange of the FR3 arm, and mount the Franka Hand parallel gripper to the FT sensor. See \figref{fig:experimental-setup}(d).

We ensure the end-effector executes slow, low-acceleration motions and remains near static equilibrium during contact, so the interaction can be regarded as quasi-static. Hence, we assume the FT sensor measures the sum of the gripper weight and the contact wrench between the object and the environment. We compensate for the gripper weight and transform the measured wrench into the end-effector frame (\figref{fig:experimental-setup}(e)) as $\gtContactWrench_{t}$, using the transformation between the FT sensor's sensing plate and the end-effector frame obtained from their mechanical drawings. When inertial effects are significant, a generalized momentum observer can be used to estimate the contact wrench between the object and the environment~\cite{luca2005sensorless}.

We assume the operational-space dynamics~\cite{khatib1987unified} and impedance control~\cite{hogan1985impedance_control,hogan2005impedance} for the system~\eqref{eqt:simulator}, with $\robotAction_{t}$ being the reference equilibrium pose. The generalized bias forces are compensated. Since during the 3 placement tasks, the robot moves in a small space, we approximate the operational-space inertia as a constant matrix.

We implement the policy~\eqref{eqt:oracle-policy} as a heuristic that interpolates from the current robot state $\robotState_{t}$ to the closest goal state in $\goalSet(\geomParam)$. We have verified that when $\geomParam = \gtGeomParam$, this policy indeed drives the robot to a goal $\goalState\in\goalSet(\gtGeomParam)$.

Our differentiable simulation does not run in real time. Since the tasks are quasi-static, we pause estimation and hold the reference pose while waiting for the next robot action. Each action takes an average of $3.36$ seconds. We expect real-time performance to be achievable with a more optimized implementation. Currently, our implementation reloads the URDF each time the geometric parameter is updated (Line~\ref{line:geometry-update} in \algref{alg:geometry-update}); enabling in-memory updates could eliminate this overhead. Simulation speed is also limited by reliance on PyTorch’s auto-differentiation, which incurs Python–C++ communication at each step~\cite{pytorch_eager, werling2021fast, yang2024jade}; combining multiple steps per call can reduce this cost. Real-time differentiable contact simulation has been demonstrated in locomotion planning~\cite{le2024fast}, and we believe similar performance is feasible for geometric uncertainty estimation.

{
\begin{algorithm}[t]
\small
  \caption{\small
    Procedure to update the belief using the history and the particle filter (PF) baseline.
  }
  \label{alg:belief-update-pf}
  \begin{algorithmic}[1]
    \Procedure{Belief-Update-PF}{
      $\belief_{t-1}$,
      $\robotState_{t - \historyLen : t}$,
      $\robotAction_{t - \historyLen : t}$,
      $\gtContactWrench_{t - \historyLen + 1 : t}$
    }
    \State
    Initialize:
    $\belief_{t} \leftarrow \varnothing$,
    $\beliefPreObs_{t} \leftarrow \varnothing$.
    \For{parameter $\geomParam$ in $\belief_{t-1}$}
    \parState
    {%
      $\geomParam \leftarrow \geomParam + \epsilon$, where $\epsilon$ is sampled from a zero-mean normal distribution.
    }
    \parState
    {%
      Compute the residual $c \leftarrow \residual_{t}(\geomParam)$ defined in~(\ref{eqt:residual-error}).
    }
    \parState
    {%
      $\beliefPreObs_{t} \leftarrow \beliefPreObs_{t} \cup \left\{\tuple{\theta, c}\right\} $.}
    \EndFor
    \parState
    {%
      $\belief_{t}\leftarrow \Call{Low-Variance-Sampler}{\beliefPreObs_{t}}$, with unnormalized weights defined in~(\ref{eqt:pf-weight}).
    }
    \State \Return $\belief_{t}$.
    \EndProcedure
  \end{algorithmic}
\end{algorithm}
}
\begingroup
\setlength{\tabcolsep}{1pt}
\begin{figure}[t]
    \centering
    \begin{tabular}{ccccc}
    \rotatebox[origin=c]{90}{\footnotesize Angle Error [deg]} &
    \includegraphics[valign=c, scale=0.2]{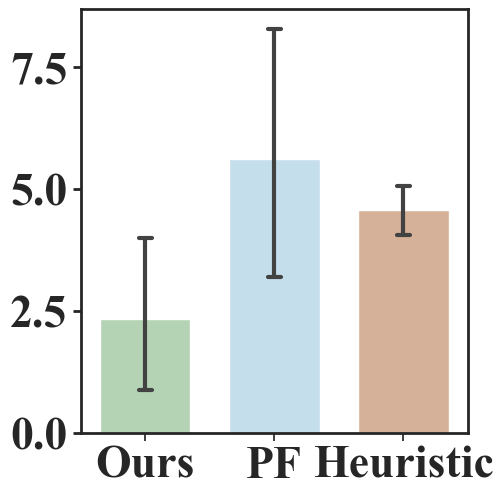} &
    \rotatebox[origin=c]{90}{\footnotesize Dist. Error [mm]} &
    \includegraphics[valign=c, scale=0.2]{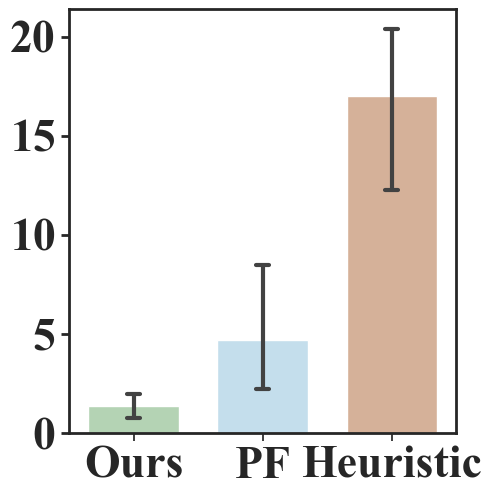} &
    \includegraphics[valign=c, scale=0.2]{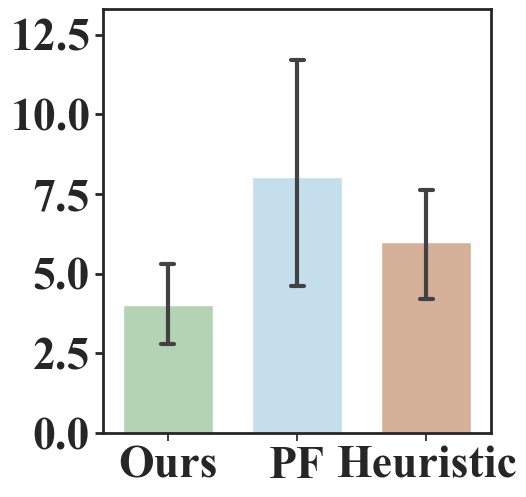} \\
    & \hspace{0.3cm}{\footnotesize (a)} & & \hspace{0.29cm}{\footnotesize (b)} & \hspace{0.32cm}{\footnotesize (c)}
    \end{tabular}
    \caption{Our method has smaller distance to the goal errors when the robot releases. We plot the means and 95\% confidence intervals over 10 test cases. For (a) \taskShape, the errors are equivalent to angle errors. For (b) \taskGrasp\ and (c) \taskPillar, the errors are equivalent to distance errors.}
    \label{fig:quantitative}
\end{figure}
\endgroup

\begingroup
\setlength{\tabcolsep}{2pt}
\begin{figure}[t]
    \centering
    \begin{tabular}{clll}
    & \multicolumn{3}{c}{
      \hspace{0.2cm}\includegraphics[valign=c, scale=0.2]{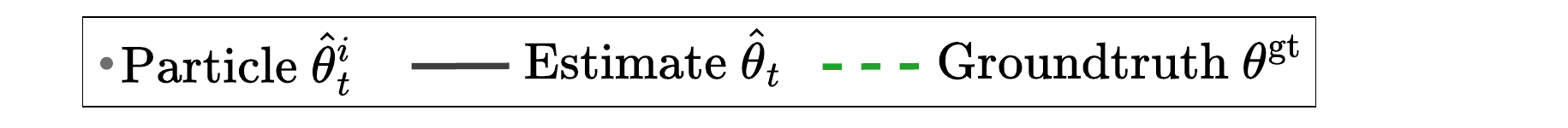}
    }\\
    & \hspace{0.8cm}{\footnotesize Ours, case \#8} &
    \hspace{0.5cm}{\footnotesize PF, case \#8} & 
    \hspace{0.5cm}{\footnotesize PF, case \#7}\\
    \rotatebox[origin=c]{90}{\footnotesize $d_{{1}}$ [mm]} &
    \includegraphics[valign=c, scale=0.2]{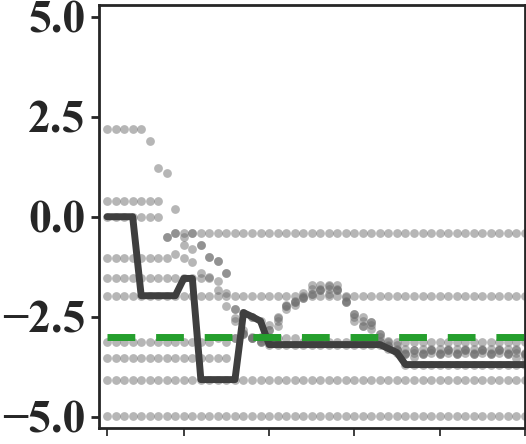} &
    \includegraphics[valign=c, scale=0.2]{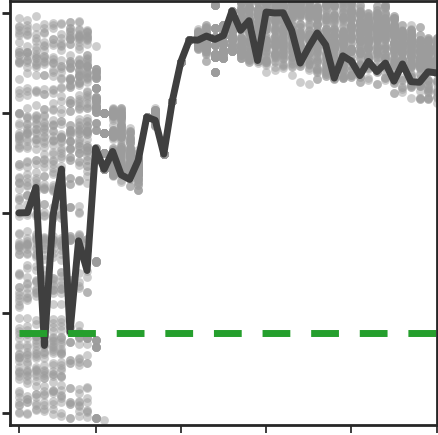} &
    \includegraphics[valign=c, scale=0.2]{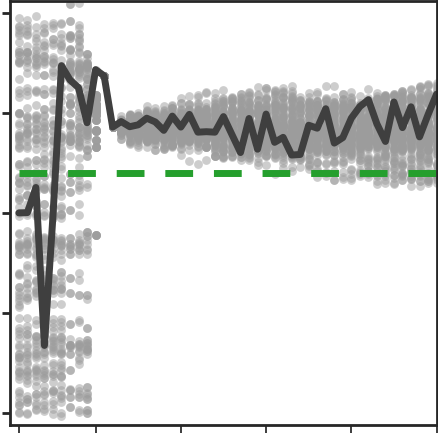} \\
    \rotatebox[origin=c]{90}{\footnotesize $d_{{2}}$ [mm]} &
    \includegraphics[valign=c, scale=0.2]{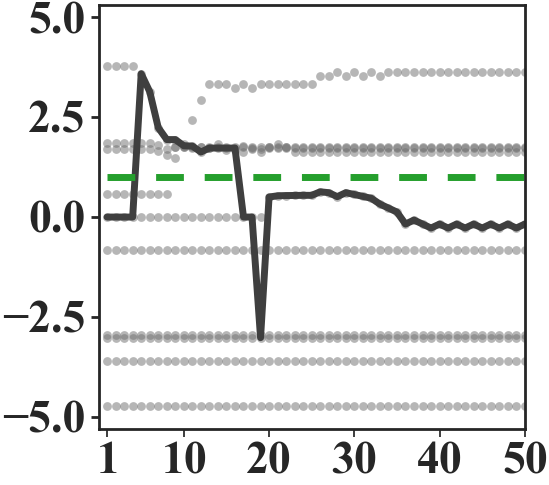} &
    \includegraphics[valign=c, scale=0.2]{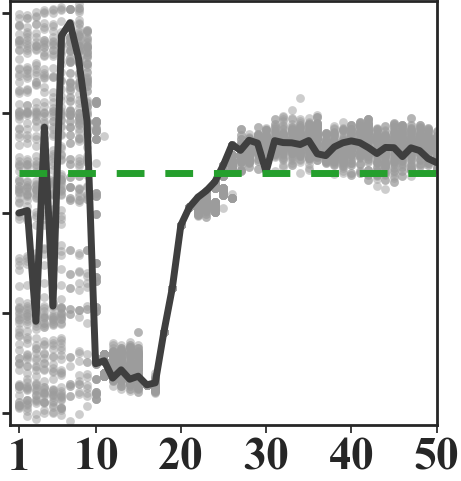} &
    \includegraphics[valign=c, scale=0.2]{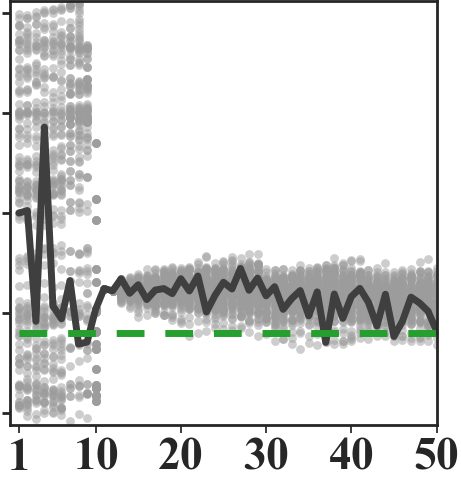} \\
    &  & \hspace{0.65cm}{\footnotesize Timestep} & \\
    & \hspace{1.4cm}{\footnotesize (a)} &
    \hspace{1.0cm}{\footnotesize (b)} &
    \hspace{1.0cm}{\footnotesize (c)}
    \end{tabular}
    \caption{PF might suffer from ``particle starvation''. We plot estimation over time from 2 test cases of the \taskShape\ task. In the same test case, ours (a) has good estimate, while PF (b) diverges on $d_{1}$ due to ``particle starvation.''  (c) In a different test case, PF maintains good estimate.}
    \label{fig:qualitative}
\end{figure}
\endgroup

\subsection{Baselines}

Granted that we can design an optimal policy for a particular task~\cite{khansari2016adaptive}, here we consider two baselines that are applicable in all scenarios. The first baseline \textbf{PF} replaces our gradient-based estimator by a particle filter~\cite{thrun2005probabilistic_robotics}. Similar to the prior works~\cite{manuelli2016localizing,sipos2022simultaneous}, we add a small Gaussian noise during the prediction update, and the unnormalized weight of the $i$-th particle is given as
\begin{equation}
\operatorname{Weight}\paren{\geomParamPar{t}{i}} = \exp\paren{- \beta \costPar{t}{i}}
\label{eqt:pf-weight}
\end{equation}
with $\beta$ being a positive constant. We implement the low variance sampler as described in~\cite{thrun2005probabilistic_robotics}. The PF implementation is summarized in \algref{alg:belief-update-pf}.

The second baseline is a trigger-based \textbf{heuristic} policy. The robot moves downwards until the measured contact wrench is larger than a threshold. This baseline demonstrates the performance of a compliant controller that does not explicitly reason about the geometric uncertainties. When there is no geometric uncertainty $\initGeomParam=\gtGeomParam$ or there is only uncertainty in heights, this heuristic policy is sufficient to reach a goal in $\goalSet(\gtGeomParam)$ for all tasks.

We use $N = 50$ particles for PF, and $N = 10$ particles in our method. This ensures that the two methods have the same number of total rollouts in each timestep.

\subsection{Evaluation Metric}

We measure the distance to goal errors when the robot releases the object. In our particular tasks, instead of measuring the distance between poses~\cite{sola2018micro,lynch2017modern}, we apply simplifications to have consistent physical dimensions. For \taskShape, since in all test cases for all methods, the cube touches the table surface, we reduce the distance to goal error to the angle between the cube's bottom surface and the table surface. For \taskGrasp\ and \taskPillar, we measure the distance between the final and goal configurations. The evaluation metrics for the 3 tasks are illustrated in \figref{fig:experimental-setup}.

\subsection{Results}

\figref{fig:quantitative} shows the errors over the 10 test cases for the 3 tasks. Our proposed method achieves smaller average errors than the PF and heuristic baselines. Interestingly, PF baseline does not always outperform the heuristic baseline on average, even though it has better best performance. When PF has bad performance, we observe ``particle starvation'', where most particles ``collapse'' to a small range of wrong values~\cite{koval2015pose}. \figref{fig:qualitative}(a)--(b) show the eighth test case from \taskShape. In \figref{fig:qualitative}(b), PF's estimate on $d_{1}$ collapsed to the wrong value, while in \figref{fig:qualitative}(a) our gradient-based estimator maintains an estimate closer to the groundtruth. The particle starvation does not always happen. \figref{fig:qualitative}(c) shows PF's estimate over time in the seventh test case from \taskShape, where PF's estimate is close to the groundtruth.

\begingroup
\setlength{\tabcolsep}{0pt}
\begin{figure}[t]
    \centering
    \begin{tabular}{cccc}
    & \multicolumn{3}{c}{
      \hspace{0.2cm}\includegraphics[valign=c, scale=0.2]{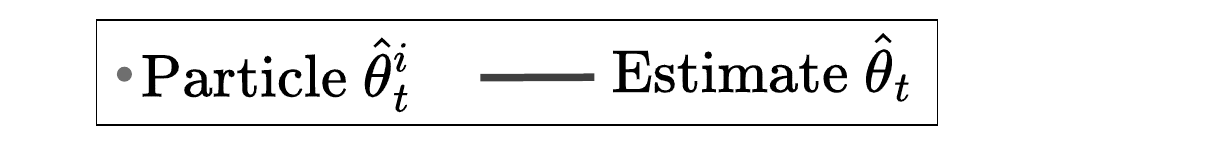}
    }\\
    &\hspace{0.2cm}{\footnotesize Lateral $l_1$} &
    \hspace{0.2cm}{\footnotesize Depth $l_2$} &
    \hspace{0.2cm}{\footnotesize Height $h$} \\
    \rotatebox[origin=c]{90}{\footnotesize Location [mm]} &
    \includegraphics[valign=c, scale=0.2]{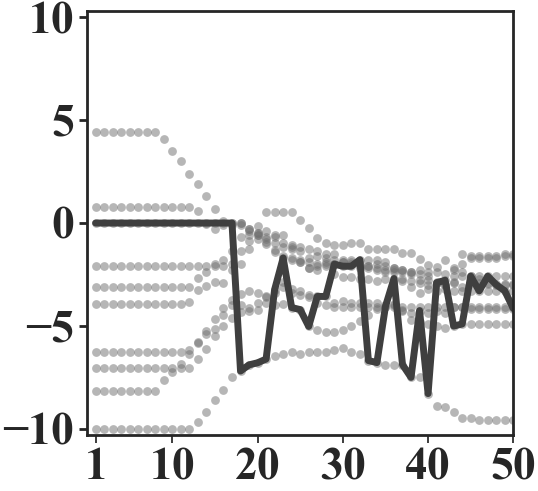} &
    \includegraphics[valign=c, scale=0.2]{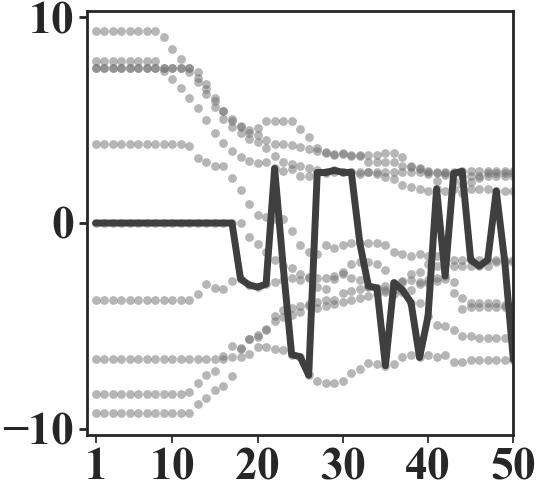} &
    \includegraphics[valign=c, scale=0.2]{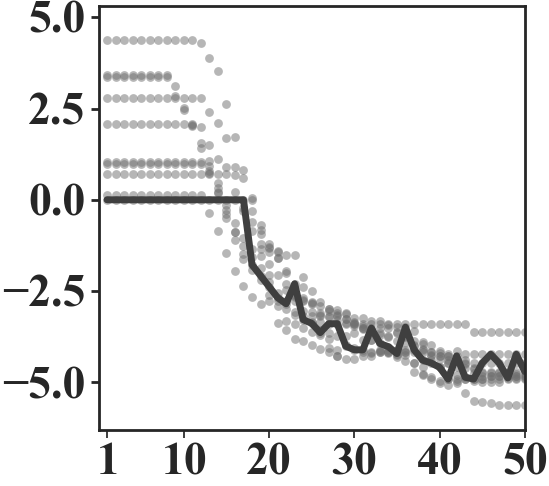}\\
    & & \hspace{0.4cm}{\footnotesize Timestep} & \\
    \end{tabular}
    \caption{Estimation of the saucer's location over time for a successful attempt. The estimate converges as the task progresses, and its variance decreases over time.}
    \label{fig:coffee-belief}
\end{figure}
\endgroup

\begingroup
\setlength{\tabcolsep}{0pt}
\begin{figure}[t]
    \centering
    \begin{tabular}{cccc}
    & \multicolumn{3}{c}{
      \hspace{0.2cm}\includegraphics[valign=c, scale=0.2]{figures/coffee/legend.pdf}
    }\\
    &\hspace{0.2cm}{\footnotesize Lateral $l_1$} &
    \hspace{0.2cm}{\footnotesize Depth $l_2$} &
    \hspace{0.2cm}{\footnotesize Height $h$} \\
    \rotatebox[origin=c]{90}{\footnotesize Location [mm]} &
    \includegraphics[valign=c, scale=0.2]{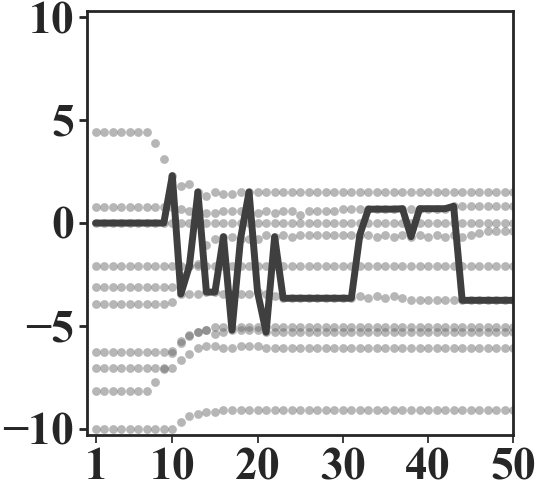} &
    \includegraphics[valign=c, scale=0.2]{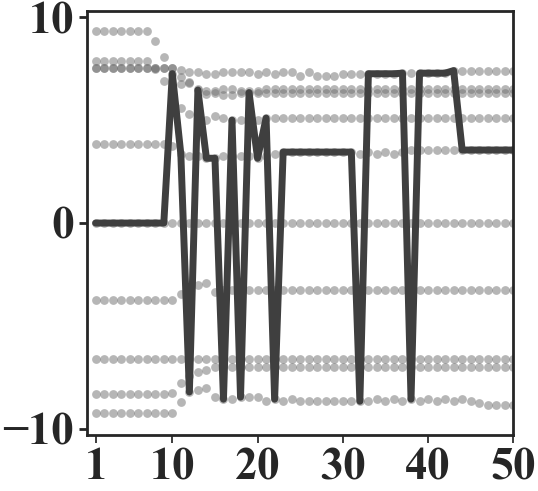} &
    \includegraphics[valign=c, scale=0.2]{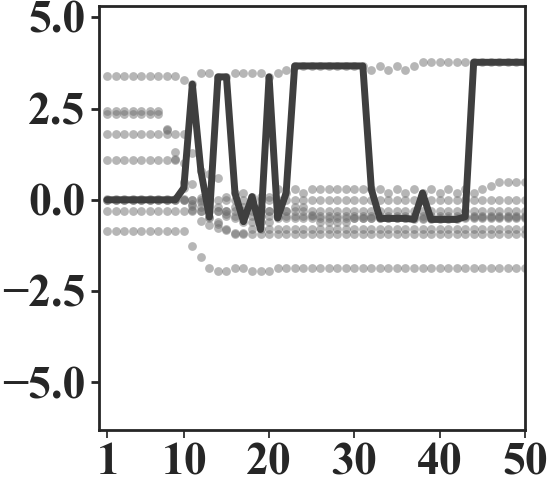}\\
    & & \hspace{0.4cm}{\footnotesize Timestep} & \\
    \end{tabular}
    \caption{Estimation of the saucer's location over time for a failed attempt. Due to poor initialization, most particles remained stagnant throughout the task, resulting in inaccurate placement.}
    \label{fig:coffee-belief-failcase}
\end{figure}
\endgroup
\begingroup
\setlength{\tabcolsep}{2pt}
\begin{figure}[t]
    \centering
    \begin{tabular}{cc}
    \includegraphics[height=58pt]{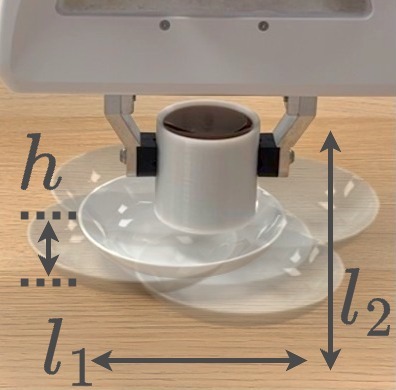} &
    \includegraphics[height=58pt]{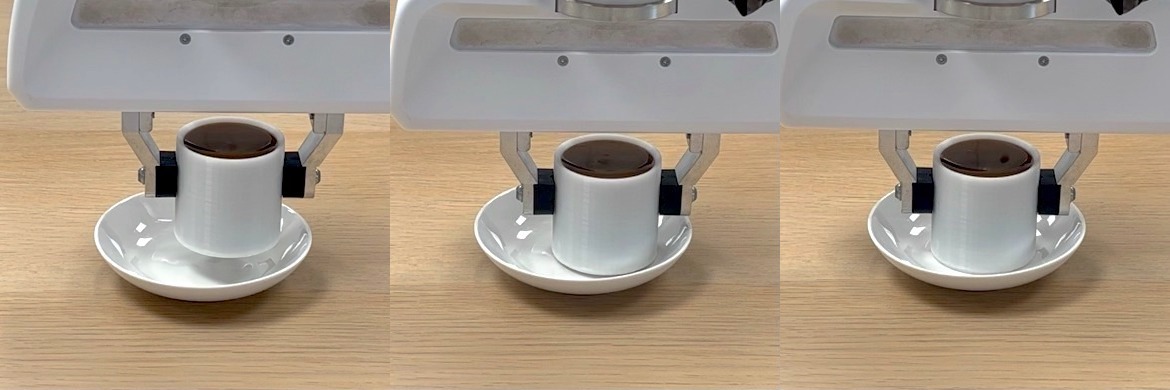} \vspace{-0.1cm} \\
    {\footnotesize (a)} & {\footnotesize (b)}
    \end{tabular}
    \caption{The coffee placement task. (a) The saucer's location, represented by $\geomParam = \left[l_1~ l_2~ h\right]^{\top}$, is uncertain. (b) During the placement process, as the estimate is refined, the robot adjusts its motion to position the cup closer to the flat center of the saucer.}
    \label{fig:coffee-setup}
\end{figure}
\endgroup

\subsection{Placing a Full Cup of Coffee}

Lastly, we deploy our method to place a full cup of coffee on a saucer (\figref{fig:coffee-setup}), relaxing several assumptions from \secref{sec:formulation}. The cup of coffee is approximated as a solid cylinder, and the saucer’s mesh is imprecise. We model the geometric uncertainty as the saucer’s 3D location, $\geomParam = [l_1~ l_2~ h]^{\top}$, shown in \figref{fig:coffee-setup}(a). Across five trials with varying saucer positions, our method successfully placed the cup without spilling in four of them, while the naive heuristic policy failed in all cases. We plot $\estGeomParam$ over time in a successful trial in \figref{fig:coffee-belief}, and a failure case in \figref{fig:coffee-belief-failcase}, where poor initialization led to stagnant particles.

\section{Conclusion and Discussion}

We presented a novel gradient that relates FT sensor readings to geometric parameters within a differentiable simulator, providing a unified approach to stable object placement under geometric uncertainty. By leveraging this gradient to minimize the discrepancy between sensor measurements and model predictions, we achieve improved performance over the gradient-free particle filter baseline. We demonstrate the effectiveness of our approach across various geometric uncertainties, including the object's in-hand pose uncertainty, the object's shape uncertainty, and the environment's shape uncertainty.

While this work focuses on geometric uncertainties in stable object placement, several directions remain open for future research. Our current approach selects robot actions based on the best estimate using an uncertainty-blind policy~\eqref{eqt:oracle-policy}, without reasoning over the full belief or actively reducing uncertainty. Future work could incorporate full belief reasoning to balance exploration and execution~\cite{thrun2005probabilistic_robotics}. Additionally, our formulation---which considers a gripper, an object, and a static environment---may extend to other contact-rich tasks under geometric uncertainty. For instance, peg-in-hole insertion can be modeled similarly, with the peg's in-hand pose and the hole's location as sources of uncertainty. Beyond geometry, future research could also address other forms of uncertainty, such as friction coefficients or sensor noise. Another improvement is to incorporate robot state predictions and measurements into the estimation to complement the force-torque data. Lastly, while our current implementation is not real-time, we believe real-time performance is achievable through further optimization, as shown in recent work on differentiable locomotion planning~\cite{le2024fast}.

\bibliographystyle{IEEEtran} 
\bibliography{references.bib}

\end{document}